\documentclass[10pt, a4paper, twocolumn]{article} % MODIFICATO: twocolumn nativo

\usepackage[utf8]{inputenc}
\usepackage[T1]{fontenc}
\usepackage[english]{babel}

\usepackage{tgheros} % Helvetica

\usepackage[varqu]{zi4} 
\usepackage{microtype}  

\usepackage{titlesec}
\titlespacing*{\section}{0pt}{1.5ex plus 1ex minus .2ex}{0.1em} 
\titlespacing*{\subsection}{0pt}{1.5ex plus 1ex minus .2ex}{0.1em}
\titlespacing*{\subsubsection}{0pt}{1.5ex plus 1ex minus .2ex}{0.1em}

\usepackage[table]{xcolor} 
\usepackage{booktabs} 
\usepackage{float}           
\usepackage{graphicx}
\usepackage{subcaption} % Mantenuto subcaption come nello scheletro originale
\usepackage[labelfont={bf,small}, textfont={small}]{caption} 

\definecolor{acqua_celeste}{RGB}{0, 160, 175}

\usepackage{amsmath}
\usepackage{amssymb}
\usepackage{mathtools}
\usepackage{amsthm}
\usepackage{bbold}     
\usepackage{pifont}     
\usepackage{multirow}
\usepackage{arydshln}   
\usepackage{array}
\usepackage{makecell}   
\usepackage{wrapfig}    
\usepackage{tabularx}  
\usepackage{bbm}

\definecolor{pgreen}{rgb}{0.13, 0.55, 0.13}
\definecolor{pred}{rgb}{0.8, 0.13, 0.13}
\definecolor{diversity}{HTML}{D4E1F5}
\definecolor{recognizability}{HTML}{FFE6CC}

\newcolumntype{Y}{>{\centering\arraybackslash}X}

\theoremstyle{plain}

\theoremstyle{definition}

\theoremstyle{remark}

\usepackage[left=1.5cm, right=1.5cm, top=1.5cm, bottom=2cm]{geometry}
\usepackage[numbers, sort&compress]{natbib}
\usepackage{hyperref}
\usepackage[capitalize,noabbrev]{cleveref} 

\hypersetup{
    colorlinks=true,        
    linkcolor=black,        
    urlcolor=acqua_celeste, 
    citecolor=acqua_celeste 
}

\usepackage{fancyhdr}
\usepackage{authblk}

\title{\huge\bfseries\vspace{-1em} Deep Variational Contrastive Learning for Joint Risk Stratification and Time-to-Event Estimation}

\author[1]{Pinar Erbil}
\author[1]{Alberto Archetti\textsuperscript{*}} 
\author[1]{Eugenio Lomurno}
\author[1]{Matteo Matteucci}

\affil[1]{Politecnico di Milano, AIRLab, Italy}

\date{}

\begin{document}

% MODIFICATO: Blocco per Titolo e Abstract a tutta larghezza
\twocolumn[
  \begin{@twocolumnfalse}
    \maketitle
    \vspace{-3em} 

    \begin{abstract}
        \setlength{\parindent}{0pt}
        \setlength{\parskip}{4pt}
        \itshape
        \noindent Survival analysis is essential for clinical decision-making, as it allows practitioners to estimate time-to-event outcomes, stratify patient risk profiles, and guide treatment planning. Deep learning has revolutionized this field with unprecedented predictive capabilities but faces a fundamental trade-off between performance and interpretability. While neural networks achieve high accuracy, their black-box nature limits clinical adoption. Conversely, deep clustering-based methods that stratify patients into interpretable risk groups typically sacrifice predictive power. We propose CONVERSE (CONtrastive Variational Ensemble for Risk Stratification and Estimation), a deep survival model that bridges this gap by unifying variational autoencoders with contrastive learning for interpretable risk stratification. CONVERSE combines variational embeddings with multiple intra- and inter-cluster contrastive losses. Self-paced learning progressively incorporates samples from easy to hard, improving training stability. The model supports cluster-specific survival heads, enabling accurate ensemble predictions. Comprehensive evaluation on four benchmark datasets demonstrates that CONVERSE achieves competitive or superior performance compared to existing deep survival methods, while maintaining meaningful patient stratification.
    \end{abstract}

    \vspace{0.5em}
    \noindent\textbf{Keywords:} Deep Survival Analysis $\cdot$ Variational Autoencoders $\cdot$ Contrastive Learning $\cdot$ Risk Stratification $\cdot$ Clustering $\cdot$ Interpretability

    \vspace{1em}
    \hrule height 1pt 
    \vspace{2em} 
  \end{@twocolumnfalse}
]

% --- REINSERIMENTO FOOTNOTE ---
{
  \renewcommand{\thefootnote}{\fnsymbol{footnote}} 
  \footnotetext[1]{Corresponding author: \texttt{alberto.archetti@polimi.it}}
}
% ------------------------------

\section{Introduction}

Survival analysis~\cite{wang2019machine} plays a fundamental role in modern healthcare, enabling practitioners to estimate time-to-event outcomes and stratify patient risk profiles. Accurate models are essential for efficient resource allocation, treatment planning, and supporting data-informed medical interventions. Beyond predictive performance, interpretability is crucial, as clinicians must trust model decisions when making safety-critical choices that directly impact patient outcomes.

Deep learning has transformed survival analysis, offering unprecedented predictive capabilities~\cite{wiegrebe2024deep}. These methods capture complex nonlinear relationships without relying on restrictive assumptions like proportional hazards~\cite{cox1972regression} and leverage high-dimensional data that traditional statistical approaches struggle to analyze. However, this enhanced performance comes at a cost: most modern survival models operate as black boxes, rendering their decision-making processes opaque. This lack of transparency poses significant challenges in clinical settings, where understanding the rationale behind predictions is as important as the predictions themselves.

To address the interpretability challenge, there has been growing interest in risk stratification approaches that integrate clustering mechanisms within survival models~\cite{chapfuwa2020survival,manduchi2021deep,nagpal2021deep,cui2024deep}. These methods discover latent subpopulations within patient cohorts, each characterized by distinct survival patterns. The appeal of stratification-based approaches lies in their inherent interpretability: patients are understood not only by their individual characteristics but also by the dynamics of their assigned risk group. Despite these advantages, clustering-based survival methods have historically lagged behind pure prediction models in terms of discriminative performance and calibration, limiting their practical adoption.

To bridge this gap, we propose CONVERSE (CONtrastive Variational Ensemble for Risk Stratification and Estimation), a deep survival model that unifies the representational power of variational autoencoders~\cite{manduchi2021deep} with the clustering capabilities of contrastive learning~\cite{cui2024deep}. Building upon a DVCSurv backbone~\cite{cui2024deep}, CONVERSE uses variational autoencoders to learn multiple-view representations of the data. These representations are then partitioned using a selection of clustering algorithms and aligned through contrastive learning, which combines both intra-cluster and inter-cluster losses. The resulting clustered representations feed into an ensemble of deep survival heads that capture population-level risk by leveraging subpopulation-specific patterns. The model is trained through a hyperparameter selection pipeline that enables the architecture to adapt to each specific dataset, incorporating only components that benefit the final model. We validate CONVERSE through comprehensive evaluation on four widely-used survival analysis datasets, demonstrating that it achieves competitive or superior performance compared to existing deep survival methods while maintaining interpretable risk stratification.

\section{Related Works}

The goal of survival analysis is to model a survival function $S(t|\mathbf{x}) = \Pr(T \geq t|\mathbf{x})$, representing the probability that an event of interest occurs after time $t$ for a subject with covariates $\mathbf{x}$. A fundamental challenge is handling censored observations, where the exact event time is unknown because the observation period ends prior to the event or the subject is lost to follow-up. While various censoring mechanisms exist, right-censoring is the most prevalent in medical applications and remains the focus of this work~\cite{wang2019machine}.

Traditional survival analysis methods fall into non-parametric, semi-parametric, and fully parametric categories. Non-parametric methods construct the survival function directly from data by averaging over population subgroups~\cite{kaplan1958nonparametric,ishwaran2008random}. Semi-parametric models combine a non-parametric baseline for the population with a parametric modulation based on patient-specific features~\cite{cox1972regression,katzman2018deepsurv}. In contrast, fully parametric models represent the survival function entirely through a finite set of parameters. Within this framework, survival functions can be defined on continuous or discrete time scales. Continuous-time formulations typically utilize classical statistical distributions with parameters estimated via nonlinear methods~\cite{nagpal2021dsm,archetti2025fpboost}. Conversely, discrete-time formulations partition the time horizon into bins $\tau_1, \tau_2, \ldots, \tau_T$. Many neural network-based approaches adopt this framework, treating survival analysis as a sequence of classification problems across time bins, where the overall survival function is reconstructed as a cumulative product~\cite{lee2018deephit,kvamme2021continuous,wiegrebe2024deep}.

To enhance interpretability, recent methods integrate clustering mechanisms within survival models to discover latent subpopulations with distinct risk profiles. Survival Cluster Analysis (SCA)~\cite{chapfuwa2020survival} employs a Bayesian non-parametric approach using a truncated Dirichlet process to automatically identify the number of clusters while constraining latent representations. VaDeSC~\cite{manduchi2021deep} utilizes a variational autoencoder framework with a Gaussian mixture prior to learn cluster-specific associations between covariates and survival times, modeling outcomes through a mixture of Weibull distributions. Deep Cox Mixtures (DCM)~\cite{nagpal2021deep} learns mixtures of Cox regression models via an EM algorithm with hard cluster assignments, estimating hazard ratios within latent clusters. More recently, DVCSurv~\cite{cui2024deep} introduced dual-view clustering through Siamese autoencoders and contrastive learning to capture robust patient representations. While these clustering-based approaches improve interpretability, they often trade off predictive performance compared to pure discriminative models.

% \section{Method}

% \begin{figure*}
%     \centering
%     \includegraphics[width=0.9\linewidth]{pipeline.jpg}
%     \caption{The overall architecture for the proposed model. Dashed lines indicate the optional paths the model can have.}
%     \label{fig:pipeline}
% \end{figure*}

% The proposed model integrates representation learning, clustering, and survival analysis into a unified pipeline as illustrated in Figure \ref{fig:pipeline}. The model is inspired by a recent work on dual-view clustering for survival analysis (DVCSurv) \cite{cui2024deep}, while maintaining a \textbf{modular design} that allows different clustering strategies and loss formulations.

% The model consists of three main components:
% \begin{enumerate}
%    \item Autoencoder(s) for learning latent representations,
%    \item Clustering module operating in the latent space(s),
%    \item Survival analysis backbone.
% \end{enumerate}

\section{Method}

\begin{figure*}
    \centering
    \includegraphics[width=\linewidth]{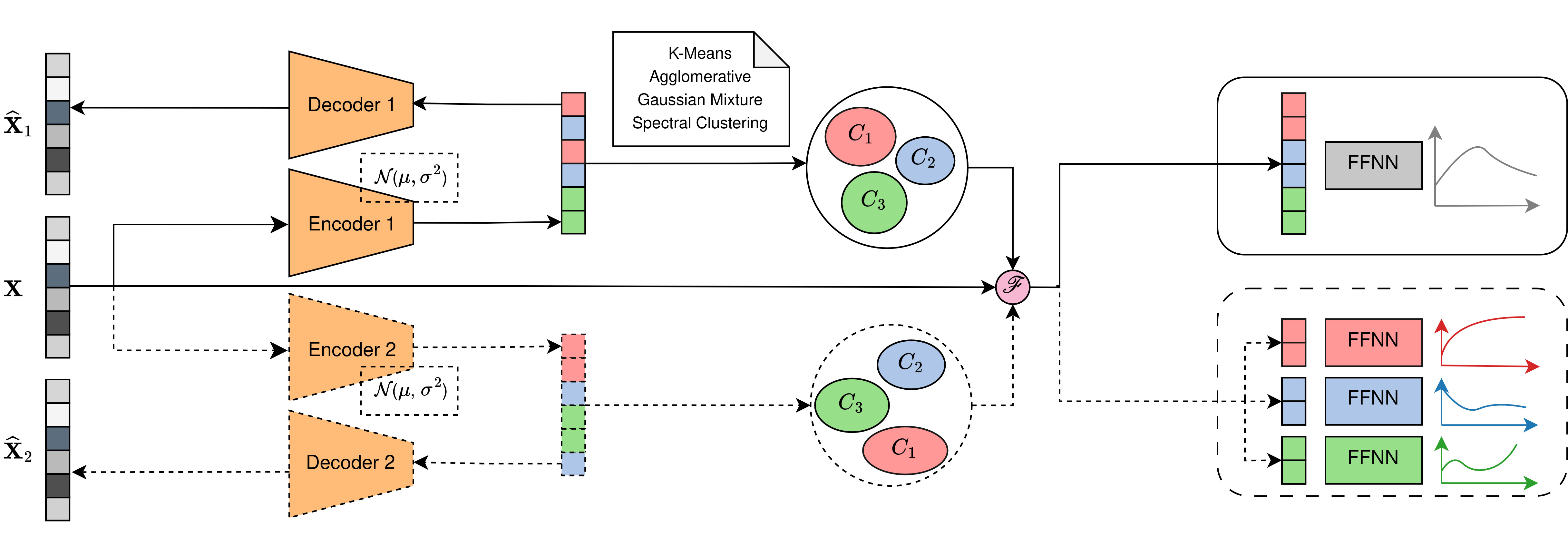}
    \vspace{-20pt}
    \caption{Overall architecture of CONVERSE. The model consists of three components: (i) variational autoencoder(s) for learning latent representations, (ii) clustering module with contrastive learning operating in the latent space, and (iii) ensemble survival prediction heads. Dashed lines indicate optional pathways depending on hyperparameter selection.}
    \vspace{-15pt}
    \label{fig:pipeline}
\end{figure*}

CONVERSE integrates representation learning, clustering, and survival analysis into a unified end-to-end trainable framework, as illustrated in Figure~\ref{fig:pipeline}. Building upon the dual-view clustering paradigm introduced in DVCSurv~\cite{cui2024deep}, CONVERSE extends this approach through a modular design that enables flexible configurations of clustering strategies, loss formulations, and survival prediction architectures.

\subsection{CONVERSE Architecture}

The model consists of three core components: (i) variational autoencoder(s) for learning latent patient representations, (ii) a clustering module enhanced with multi-view contrastive learning, and (iii) an ensemble of survival prediction heads that can be either shared or cluster-specific.

% Throughout this section, we use the following notation: $\mathbf{x}_i \in \mathbb{R}^D$ denotes the input covariates for patient $i$, $t_i$ represents the observed time, and $e_i \in \{0,1\}$ is the event indicator where $e_i = 1$ indicates an observed event and $e_i = 0$ indicates censoring. The total number of patients is $N$, and we discretize the time horizon into $T$ bins.

\subsubsection{Representation Learning and Autoencoders}

The representation learning component learns a low-dimensional latent space that captures the essential structure of patient covariates. This latent space serves as the foundation for both clustering and survival prediction. Differently from~\cite{cui2024deep}, we employ variational autoencoders (VAEs)~\cite{manduchi2021deep} to enable more flexible and regularized representations compared to standard autoencoders. Given an input vector $\mathbf{x}_i$, a variational encoder maps it to the parameters of a Gaussian distribution $q_\phi(\mathbf{z}_i|\mathbf{x}_i) = \mathcal{N}(\boldsymbol{\mu}_i, \text{diag}(\boldsymbol{\sigma}_i^2))$, where $\boldsymbol{\mu}_i, \boldsymbol{\sigma}_i \in \mathbb{R}^d$ are the mean and standard deviation vectors produced by the encoder network. A decoder network then reconstructs the input from the latent representation $\mathbf{z}_i \in \mathbb{R}^d$. The reconstruction loss encourages the learned representation to preserve information from the original covariates:
\begin{equation}
    \mathcal{L}_{\text{REC}} = \frac{1}{N}\sum_{i=1}^N \| \hat{\mathbf{x}}_i - \mathbf{x}_i \|_2^2,
\label{eq:loss_rec}
\end{equation}
where $\hat{\mathbf{x}}_i$ is the reconstructed input. To regularize the latent space and prevent overfitting, we impose a KL divergence constraint that encourages the approximate posterior to remain close to a standard Gaussian prior $p(\mathbf{z}) = \mathcal{N}(\mathbf{0}, \mathbf{I})$:
\begin{equation}
        \mathcal{L}_{\text{KLD}} = \frac{1}{N}\sum_{i=1}^N \text{KL}\big(q_\phi(\mathbf{z}_i \mid \mathbf{x}_i) \,\|\, p(\mathbf{z})\big).
\label{eq:loss_kld}
\end{equation}
The framework also supports standard (non-variational) autoencoders by omitting the KL divergence term, in which case the encoder directly produces a deterministic latent representation $\mathbf{z}_i$ without sampling. The choice between variational or deterministic autoencoder is handled by a hyperparameter optimization procedure per dataset.

To capture complementary views of the patient data as in~\cite{cui2024deep}, CONVERSE optionally employs Siamese autoencoders. This configuration uses two encoders with identical architectures but independent parameters, creating two distinct latent representations $\mathbf{z}_i^{(1)}, \mathbf{z}_i^{(2)} \in \mathbb{R}^d$ from the same input $\mathbf{x}_i$. In the Siamese configuration, the reconstruction and KL divergence losses are averaged across the two views.

\subsubsection{Clustering and Contrastive Learning}

After learning latent representations, CONVERSE discovers patient subpopulations through clustering in the latent space and refines these representations through contrastive learning, following the steps from DSACC~\cite{cui2024dsac} and DVCSurv~\cite{cui2024deep}. Clustering is performed directly in the learned latent space(s). For a given latent representation (either $\mathbf{z}_i$ in the single encoder case or $\mathbf{z}_i^{(v)}$ for view $v$ in the Siamese case), $K$ cluster centers are denoted by $\mathbf{M} = [\mathbf{m}_1, \ldots, \mathbf{m}_K]$ where $\mathbf{m}_k \in \mathbb{R}^d$. Each patient is assigned to exactly one cluster, represented by a one-hot vector $\mathbf{c}_i \in \{0,1\}^K$ where $c_{i,k} = 1$ if patient $i$ belongs to cluster $k$. The clustering loss minimizes the distance between each latent representation and its assigned cluster center as
\begin{equation}
    \mathcal{L}_{\text{CLUS}} = \frac{1}{N}\sum_{i=1}^N \| \mathbf{z}_i - \mathbf{M}\mathbf{c}_i \|_2^2.
\label{eq:loss_clus}
\end{equation}
For the Siamese configuration, clustering is performed independently in each view, and the loss is averaged:
\begin{equation}
    \mathcal{L}_{\text{CLUS}}^{\text{Siam}} = \frac{1}{2N}\sum_{i=1}^N \sum_{v=1}^2 \| \mathbf{z}_i^{(v)} - \mathbf{M}^{(v)}\mathbf{c}_i^{(v)} \|_2^2.
\label{eq:loss_clus_siam}
\end{equation}
Differently from~\cite{cui2024dsac,cui2024deep}, our framework supports multiple clustering algorithms including $K$-means, agglomerative clustering, Gaussian mixture models, and spectral clustering. The specific clustering choice is handled by the hyperparameter optimization procedure.

Following~\cite{cui2024deep}, contrastive learning further refines the learned representations by encouraging similar patients to cluster together while separating dissimilar ones. Up to three complementary contrastive objectives are employed, depending on the encoder configuration. The first objective, Intra-View Cluster-Guided contrastive learning (IVCG), leverages information from uncensored patients to improve representations of censored patients. For each censored patient (anchor), positive pairs are formed with uncensored patients in the same cluster and negative pairs with uncensored patients in different clusters. This design is motivated by the observation that patients in the same risk group should have similar latent representations, regardless of censoring status. Using the InfoNCE loss, this is formalized as:
\begin{equation}
\mathcal{L}_{\text{IVCG}} = -\frac{1}{N_{\text{cens}}} \sum_{\substack{i=1 \\ e_i = 0}}^{N} \sum_{\substack{j \in \mathcal{P}_i^+}}
\log \frac{\exp\!\left( s(\mathbf{z}_i, \mathbf{z}_j) / \tau \right)}
{\sum_{k=1}^{N} \exp\!\left( s(\mathbf{z}_i, \mathbf{z}_k) / \tau \right)},
\label{eq:loss_ivcg}
\end{equation}
where $N_{\text{cens}} = \sum_{i=1}^N (1-e_i)$ is the number of censored patients, $\mathcal{P}_i^+ = \{j : e_j = 1, c_j = c_i\}$ is the set of uncensored patients in the same cluster as patient $i$, $s(\cdot, \cdot)$ denotes cosine similarity, and $\tau$ is a temperature hyperparameter. For Siamese encoders, we sum the losses from both views.

The second objective, Inter-View Instance-Wise contrastive learning (IVIW), is specific for Siamese encoders and its goal is to encourage consistency by treating the two latent representations of the same patient as positive pairs, while all other patients form negative pairs as
\begin{equation}
    \mathcal{L}_{\text{IVIW}} = -\frac{1}{N} \sum_{i=1}^{N} \sum_{v=1}^2 
    \log \frac{\exp\!\left( s(\mathbf{z}_i^{(v)}, \mathbf{z}_i^{(v')}) / \tau \right)}
    {\sum_{j=1}^N \exp\!\left( s(\mathbf{z}_i^{(v)}, \mathbf{z}_j^{(v')}) / \tau \right)}.
\label{eq:loss_ivig}
\end{equation}

Lastly, the third objective, Inter-View Cluster-Wise contrastive learning (IVCW), is introduced to enforce consistency between the two latent views. In addition to the hard cluster assignments produced by the clustering algorithm, soft cluster assignments are incorporated using Student's t-distribution to capture the clustering confidence. The goal is to get, for each instance $\mathbf{z}_i^{(v)}$, how confidently it belongs to the cluster $k$ in view $v$. This confidence is computed as:
\begin{equation}
    q_{i,k}^{(v)} =
    \frac{\left( 1 + \left\lVert \mathbf{z}_{i}^{(v)} - \mathbf{m}_{k}^{(v)} \right\rVert_{2}^{2} / \nu \right)^{-\frac{\nu + 1}{2}}}
    {\sum_{k'=1}^K \left( 1 + \left\lVert \mathbf{z}_{i}^{(v)} - \mathbf{m}_{k'}^{(v)} \right\rVert_{2}^{2} / \nu \right)^{-\frac{\nu + 1}{2}}}.
\label{eq:soft_assignment}
\end{equation}
where $\nu$ represents the degrees of freedom. For a fixed cluster $k$ and view $v$, the confidence scores across all patients are stacked into $\mathbf{q}_k^{(v)} = [q_{1,k}^{(v)}, \ldots, q_{N,k}^{(v)}]^\top \in \mathbb{R}^N$, which represents the cluster-level distribution in that view. The cluster-wise contrastive loss encourages corresponding clusters across views to have similar distributions:
\begin{equation}
    \mathcal{L}_{\text{IVCW}} = - \frac{1}{2K} \sum_{k=1}^{K} \sum_{v=1}^2 \log\frac
    {\exp\!\bigl( s(\mathbf{q}_{k}^{(v)}, \mathbf{q}_{k}^{(v')}) / \tau \bigr)}
    {\sum_{k'=1}^{K} \exp\!\bigl( s(\mathbf{q}_{k}^{(v)}, \mathbf{q}_{k'}^{(v')}) / \tau\bigr)}.
\label{eq:loss_ivcw}
\end{equation}
The total contrastive loss combines all applicable objectives with learnable weights:
\begin{equation}
    \mathcal{L}_{\text{CL}} = \alpha_{\text{IVCG}}\mathcal{L}_{\text{IVCG}} + \alpha_{\text{IVIG}}\mathcal{L}_{\text{IVIG}} + \alpha_{\text{IVCW}}\mathcal{L}_{\text{IVCW}},
\label{eq:loss_cl_total}
\end{equation}
where $\alpha_{\text{IVIG}}$ and $\alpha_{\text{IVCW}}$ are set to zero when using a single encoder.

\subsubsection{Ensemble Survival Prediction}

The final component maps the learned latent representations to survival predictions. We adopt a discrete-time formulation based on an ensemble of feed-forward neural networks that predicts the probability of the event occurring in each time bin $\tau_1, \ldots, \tau_T$. We have two architectural variants for this component: shared-head and cluster-specific-heads. In the shared-head configuration, all patients are processed by a single survival network. The input combines the latent representation(s) with the original covariates to preserve both learned structure and raw clinical information:
\begin{equation}
\mathbf{h}_i =
\begin{cases}
    [\mathbf{z}_i; \mathbf{x}_i] & \text{if single encoder}, \\[4pt]
    \left[\frac{1}{2}(\mathbf{z}_i^{(1)} + \mathbf{z}_i^{(2)}); \mathbf{x}_i\right] & \text{if dual encoder},
\end{cases}
\label{eq:survival_input}
\end{equation}
A single feed-forward neural network $g_{\text{surv}}(\cdot)$ processes this combined representation to output discrete-time event probabilities $\{\hat{p}_{i,t}\}_{t=1}^{T} = g_{\text{surv}}(\mathbf{h}_i)$, where $\hat{p}_{i,t}$ represents the predicted probability that the event occurs at time bin $t$ for patient $i$.

Alternatively, in the cluster-specific-heads configuration, we employ an ensemble of survival heads $\{g_{\text{surv}}^{(k)}\}_{k=1}^K$, where each cluster $k$ is assigned to its own head. This design is motivated by the hypothesis that different patient subpopulations may exhibit fundamentally different survival dynamics that are better captured by specialized models. For a patient $i$ assigned to cluster $k$, the prediction is $\{\hat{p}_{i,t}\}_{t=1}^{T} = g_{\text{surv}}^{(k)}(\mathbf{h}_i)$, where the input $\mathbf{h}_i$ is constructed as in Equation~\eqref{eq:survival_input}. This ensemble approach allows the model to learn cluster-specific risk patterns while maintaining a shared latent representation.

We optimize the survival heads using two complementary objectives. The negative log-likelihood (NLL) loss ensures well-calibrated probability estimates:
\begin{equation}
    \mathcal{L}_{\text{NLL}} = -\frac{1}{N} \sum_{i=1}^{N} \left[ e_i \log(\hat{p}_{i,t_i}) + (1 - e_i) \log(\hat{S}_{i,t_i}) \right].
\label{eq:loss_nll}
\end{equation}
To improve discriminative performance, a ranking loss promotes the correct ordering of survival probabilities for comparable pairs. Two patients $(i,j)$ are comparable if patient $i$ experiences the event before patient $j$ and is uncensored ($e_i = 1$ and $t_i < t_j$). The ranking loss is defined as
\begin{equation}
    \mathcal{L}_{\text{RANK}} = \sum_{\substack{i,j \\ e_i = 1, t_i < t_j}} \exp\left( \frac{\hat{S}_{i,t_i} - \hat{S}_{j,t_i}}{\sigma} \right),
\label{eq:loss_rank}
\end{equation}
where $\sigma$ is a hyperparameter. The combined survival loss is $\mathcal{L}_{\text{SURV}} = \mathcal{L}_{\text{NLL}} + \beta \mathcal{L}_{\text{RANK}}$, where $\beta$ balances calibration and discrimination.

\subsection{Training Procedure}

CONVERSE is trained in three stages: pre-training, cluster initialization, and end-to-end refinement with self-paced learning. Self-paced learning (SPL)~\cite{kumar2010self} progressively incorporates samples from easy to hard during optimization, improving training stability. Easy instances are those on which the model exhibits high confidence (low loss), while hard instances lie near cluster boundaries where assignments are ambiguous. Following~\cite{cui2024deep}, SPL is applied only to the representation learning and clustering components. At each training epoch $e$, an adaptive threshold $\lambda_e$ is computed as
\begin{equation}
    \lambda_e = \mu(\{\mathcal{L}_i\}_{i=1}^N) + \frac{e}{E_{\text{max}}}\sigma(\{\mathcal{L}_i\}_{i=1}^N),
\end{equation}
where $E_{\text{max}}$ is the total number of epochs, and $\mu(\cdot)$ and $\sigma(\cdot)$ denote the mean and standard deviation of the instance-level losses. Each instance-level loss combines reconstruction, KL divergence, and clustering terms:
\begin{equation}
    \mathcal{L}_i = \alpha_{\text{REC}}\mathcal{L}_{\text{REC},i} + \alpha_{\text{KLD}}\mathcal{L}_{\text{KLD},i} + \alpha_{\text{CLUS}}\mathcal{L}_{\text{CLUS},i},
\end{equation}
where $\alpha_{\text{REC}}, \alpha_{\text{KLD}}, \alpha_{\text{CLUS}}$ are weighting hyperparameters, and $\mathcal{L}_{\text{KLD},i}$ is omitted when using non-variational encoders. Only instances with $\mathcal{L}_i \leq \lambda_e$ contribute to the SPL objective:
\begin{equation}
    \mathcal{L}_{\text{SPL}} = \frac{1}{\sum_{i=1}^N \mathbbm{1}({\mathcal{L}_i \leq \lambda_e})} \sum_{i=1}^N \mathbbm{1}({\mathcal{L}_i \leq \lambda_e)} \cdot \mathcal{L}_i.
\end{equation}
As training progresses, $\lambda_e$ increases linearly, gradually admitting harder instances into the optimization.

The first stage performs pre-training of the autoencoder(s) and survival head(s) jointly by minimizing
\begin{equation}
    \min_{\Theta} \left( \alpha_{\text{REC}}\mathcal{L}_{\text{REC}} + \alpha_{\text{KLD}}\mathcal{L}_{\text{KLD}} + \alpha_{\text{SURV}}\mathcal{L}_{\text{SURV}} \right),
\label{eq:pretrain}
\end{equation}
where $\Theta$ denotes all network parameters. This stage initializes meaningful representations before introducing clustering.

In the second stage, we apply the selected clustering algorithm (e.g., $K$-means) to the learned latent representations to obtain initial cluster centers $\mathbf{M}$ and assignments $\mathbf{c}_i$. These centers remain fixed throughout the subsequent training stage.

Finally, the third stage trains the full model end-to-end while enabling self-paced learning by optimizing
\begin{equation}
    \min_{\Theta} \left( \alpha_{\text{SPL}}\mathcal{L}_{\text{SPL}} + \alpha_{\text{CL}}\mathcal{L}_{\text{CL}} + \alpha_{\text{SURV}}\mathcal{L}_{\text{SURV}} \right).
\label{eq:endtoend}
\end{equation}
After each epoch, we update the cluster assignments $\mathbf{c}_i$ for all patients by reassigning each to its nearest cluster center. All weighting hyperparameters ($\alpha_{\cdot}, \beta$), the temperature $\tau$, and architectural choices are jointly optimized via the validation procedure described in Section~\ref{subsec:optimization}.

\section{Experiments}

\subsection{Datasets and Baseline Models}

We evaluate our model on four widely used survival analysis benchmarks. Three datasets focus on breast cancer outcomes: GBSG (recurrence-free survival), METABRIC, and TCGA\_BRCA (overall survival). The fourth dataset, WHAS, evaluates survival following myocardial infarction~\cite{polsterl2020scikit,katzman2018deepsurv,ogier2022flamby}. For performance comparison, we consider both standard neural network-based methods and clustering-based approaches. The former includes DeepSurv~\cite{katzman2018deepsurv} and DeepHit~\cite{lee2018deephit}, which serve as discriminative baselines. The latter comprises DCM~\cite{nagpal2021deep}, SCA~\cite{chapfuwa2020survival}, VADESC~\cite{manduchi2021deep}, and DVCSurv~\cite{cui2024deep}.

\subsection{Experimental Procedure and Hyperparameter Optimization}
\label{subsec:optimization}
To ensure reliable out-of-sample evaluation on these small-scale survival datasets, we employ a rigorous validation procedure for all models. We use Optuna~\cite{akiba2019optuna} with a Tree-structured Parzen Estimator to sample hyperparameter configurations. For each configuration, we generate five bootstrap splits of the original data, creating five independent training (60\%), validation (20\%), and test (20\%) sets. Each split is independently preprocessed using normalization and one-hot encoding for categorical variables. We train each model on all five splits separately and select the best hyperparameter configuration based on validation performance, jointly considering both concordance index (C-Index) and integrated Brier score (IBS). Final test results are reported for the configuration achieving the highest average validation performance averaged across the five splits. Survival estimations on discrete time grids from all models are linearly interpolated~\cite{archetti2024bridging}.

\subsection{Results}

\begin{table}[t]
\centering
\caption{C-Index Test Results ($\uparrow$) Across Datasets.}
\vspace{-10pt}
\label{table:cindex}
\begin{center}
\begin{footnotesize}
\begin{sc}
\begin{tabularx}{\linewidth}{@{} l @{} Y @{} Y @{} Y @{} c @{}}
\toprule
Model & GBSG & METABRIC & WHAS & TCGA\_BRCA \\
\midrule
DeepSurv & $67.2 \pm 4.0$ & $64.6 \pm 1.9$ & \underline{$77.4 \pm 4.7$} & $\mathbf{74.3 \pm 4.7}$ \\
DeepHit & $66.6 \pm 3.0$ & $64.8 \pm 1.8$ & $77.3 \pm 4.1$ & $72.3 \pm 4.4$ \\
VADESC & \underline{$67.3 \pm 2.8$} & $64.5 \pm 2.2$ & $77.3 \pm 4.0$ & \underline{$72.5 \pm 5.8$} \\
DCM & $65.9 \pm 2.4$ & $64.5 \pm 2.1$ & $76.4 \pm 4.2$ & $70.7 \pm 5.1$ \\
SCA & $66.6 \pm 3.2$ & $64.7 \pm 2.2$ & $75.2 \pm 3.6$ & $67.8 \pm 8.4$ \\
DVCSurv & $\mathbf{67.8 \pm 2.7}$ & \underline{$65.8 \pm 1.7$} & $76.4 \pm 4.4$ & $68.2 \pm 6.6$ \\
\midrule
CONVERSE & $\mathbf{67.8 \pm 2.5}$ & $\mathbf{66.0 \pm 1.8}$ & $\mathbf{77.6 \pm 3.8}$ & $69.3 \pm 4.9$ \\
\bottomrule
\end{tabularx}
\end{sc}
\end{footnotesize}
\end{center}
\end{table}

\begin{table}[t]
\centering
\caption{IBS Test Results ($\downarrow$) Across Datasets.}
\vspace{-10pt}
\label{table:ibs}
\begin{center}
\begin{footnotesize}
\begin{sc}
\begin{tabularx}{\linewidth}{@{} l @{} Y @{} Y @{} Y @{} c @{}}
\toprule
Model & GBSG & METABRIC & WHAS & TCGA\_BRCA \\
\midrule
DeepSurv & \underline{$18.7 \pm 1.3$} & \underline{$19.5 \pm 0.8$} & $\mathbf{15.4 \pm 2.6}$ & $\mathbf{10.2 \pm 1.1}$ \\
DeepHit & $19.2 \pm 0.9$ & $\mathbf{19.4 \pm 0.7}$ & \underline{$15.6 \pm 2.4$} & \underline{$10.5 \pm 0.7$} \\
VADESC & $19.3 \pm 0.9$ & $\mathbf{19.4 \pm 0.8}$ & $16.8 \pm 2.3$ & $16.2 \pm 2.1$ \\
DCM & $19.4 \pm 1.1$ & $\mathbf{19.4 \pm 0.8}$ & $16.7 \pm 2.5$ & $11.1 \pm 1.0$ \\
SCA & $31.7 \pm 3.3$ & $27.4 \pm 1.5$ & $23.9 \pm 6.1$ & $27.8 \pm 7.6$ \\
DVCSurv & $20.4 \pm 1.1$ & $22.7 \pm 0.8$ & $26.8 \pm 7.8$ & $13.3 \pm 2.2$ \\
\midrule
CONVERSE & $\mathbf{18.6 \pm 1.3}$ & $21.1 \pm 0.8$ & $16.1 \pm 2.4$ & $11.5 \pm 0.9$ \\
\bottomrule
\end{tabularx}
\end{sc}
\end{footnotesize}
\end{center}
\end{table}

We evaluate CONVERSE against two categories of baseline methods: neural-based models (DeepSurv and DeepHit), which prioritize predictive performance without explicit risk stratification, and cluster-based models (VADESC, DCM, SCA, and DVCSurv), which integrate clustering for patient subgroups. Tables~\ref{table:cindex} and~\ref{table:ibs} report C-Index and IBS results, respectively, with the best performance in bold and second-best underlined. Values are scaled by a factor of 100 to improve readability.

CONVERSE achieves the highest C-Index on three of four datasets (GBSG, METABRIC, and WHAS), demonstrating strong discriminative performance. Compared to cluster-based methods, CONVERSE shows a consistent average improvement of $+0.70\%$. This suggests that the proposed integration of variational autoencoders effectively enhances the quality of learned representations for survival prediction. Most notably, CONVERSE equals or outperforms DVCSurv on all datasets. Against neural baselines, CONVERSE achieves comparable but not superior performance on average.

\noindent Results on IBS reveal complementary insights. CONVERSE achieves the best calibration on GBSG ($18.6$) and demonstrates competitive performance on WHAS and TCGA\_BRCA. Compared to cluster-based methods, CONVERSE improves calibration substantially, averaging $+1.63\%$ in IBS over clustering-based approaches, indicating that ensemble survival heads can refine probability estimates effectively. At the same time, CONVERSE exhibits a slight degradation in calibration with respect to neural models on the METABRIC dataset. This reflects a fundamental trade-off inherent to representation-based models: partitioning the feature space through clustering can introduce additional uncertainty in probability estimates, particularly for patients near cluster boundaries. Nevertheless, the calibration gap remains modest, and CONVERSE substantially narrows the performance difference between neural and cluster-based approaches compared to prior work.

\section{Interpretability}
\begin{figure}[t]
    \centering
    \includegraphics[width=\linewidth]{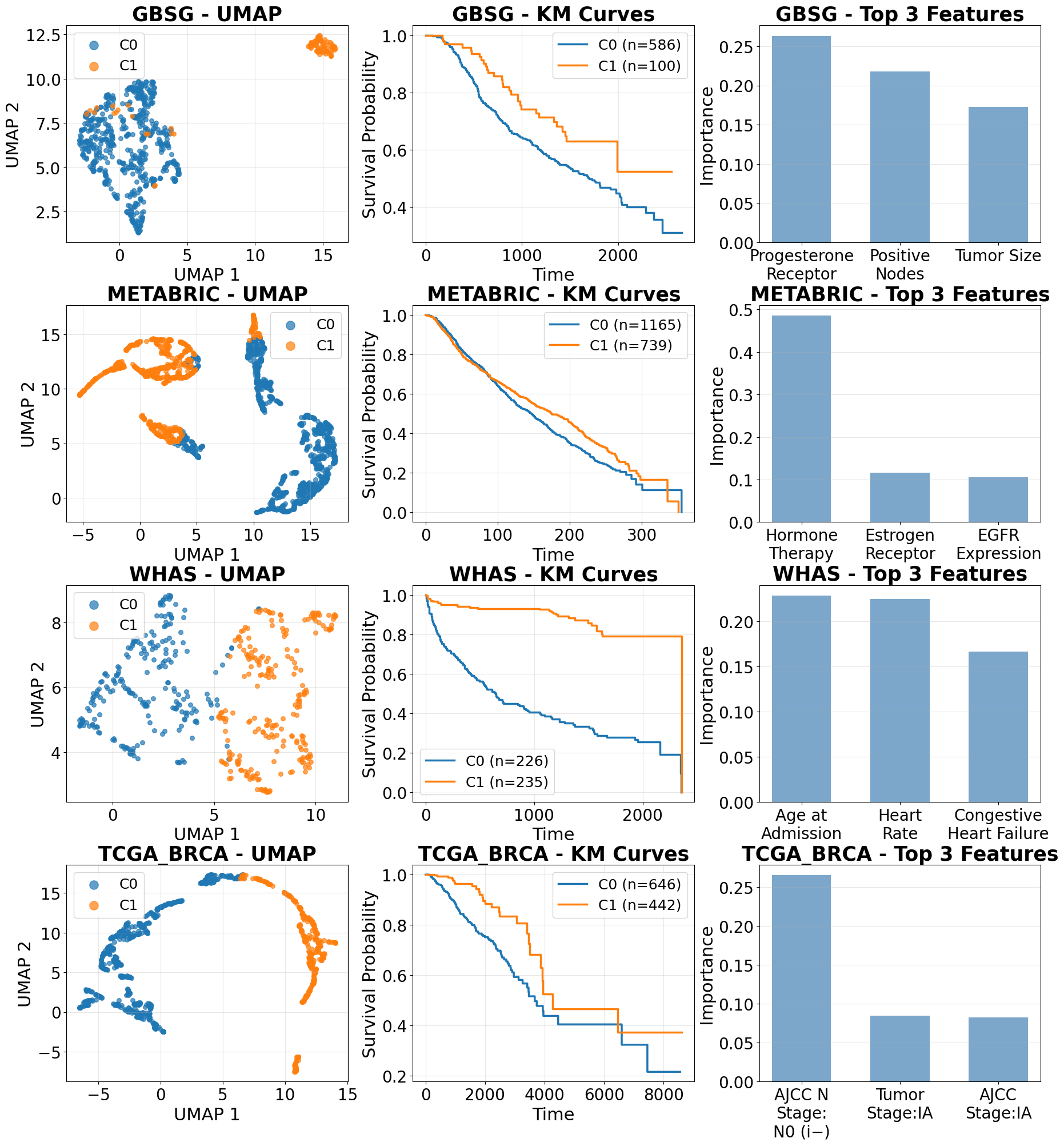}
    \vspace{-5pt}
    \caption{Representation visualizations from trained CONVERSE architectures. The first column collects UMAP visualizations of the latent spaces with $K$-means cluster assignments; the second column shows Kaplan--Meier curves for the resulting clusters; the third column presents Gini-based feature importance to highlight the covariates driving each group.}
    \vspace{-10pt}
    \label{fig:explainability}
\end{figure}

\noindent To demonstrate that CONVERSE learns clinically meaningful patient representations, we present a set of visualizations in Figure~\ref{fig:explainability} taken from models including the best-performing hyperparameters for each dataset. The figure is organized into three columns highlighting different aspects of the learned representations. The first column shows UMAP 2D projections~\cite{mcinnes2018umap} of the latent space for all four datasets, colored by cluster assignments from $K$-Means clustering with $k=2$. We fix the number of clusters to two to investigate the extent to which the latent space naturally separates patients into higher-risk and lower-risk groups. Across all datasets, the clusters are well-separated in the latent space, indicating that CONVERSE learns structured and discriminative representations rather than arbitrary partitions.

\noindent The second column presents Kaplan--Meier survival curves for the two clusters in each dataset. All four datasets exhibit clear separation between clusters, with statistically significant differences confirmed by log-rank p-values ($p < 0.05$). Consistently, one cluster exhibits higher risk while the other shows more favorable survival profiles (low-risk). This separation demonstrates that the latent representation captures clinically relevant risk information.

\noindent Finally, the third column summarizes cluster-level feature importance. To obtain these estimations, we trained a Random Forest classifier using cluster indices as labels and computed Gini-based~\cite{breiman2017classification} feature importances to identify which covariates most strongly differentiate the learned subpopulations. For the breast cancer datasets, the top features align with established prognostic factors. In GBSG, which measures recurrence-free survival, the most influential features are progesterone receptor status, number of positive lymph nodes, and tumor size—all well-established predictors of disease recurrence~\cite{early2011relevance}. In METABRIC, hormone therapy dominates feature importance, followed by estrogen receptor status and EGFR expression, reflecting the substantial effect of treatment response on overall survival in this cohort. In TCGA\_BRCA, the top features are all related to cancer staging, which is clinically appropriate as stage remains one of the most powerful prognostic factors for overall survival in breast cancer. For WHAS, which focuses on survival following myocardial infarction, the highest-ranked features are age at admission, heart rate, and congestive heart failure, all falling within the major risk factors for cardiovascular mortality~\cite{arnett20192019}. These results demonstrate that CONVERSE identifies meaningful subpopulations characterized by features that align with clinical knowledge for each specific outcome and dataset.

% \begin{figure*}[t]
%     \centering
%     \includegraphics[width=0.75\linewidth]{explainability2.png}
%     \caption{Explainability results for the proposed model. The figure summarizes the discovered risk groups by combining (i) a UMAP visualization of the learned latent space with K-means cluster assignments, (ii) Kaplan-Meier curves for the resulting clusters, and (iii) SHAP-based feature importance to highlight the covariates driving each group.}
%     \label{fig:explainability2}
% \end{figure*}

\section{Conclusion}

\noindent We presented CONVERSE, a deep survival model that bridges the gap between discriminative performance and interpretable risk stratification through the integration of variational autoencoders, multi-view contrastive learning, and cluster-specific survival heads. Comprehensive evaluation on four datasets demonstrates that CONVERSE achieves competitive performance compared to neural network approaches while outperforming cluster-based methods. Additionally, CONVERSE provides clinically interpretable patient stratification, with clusters corresponding to well-established prognostic factors, providing a practical solution for data-informed clinical decision-making.

\section*{Acknowledgements}
This paper is supported by the FAIR (Future Artificial Intelligence Research) project, funded by the NextGenerationEU program within the PNRR-PE-AI scheme (M4C2, investment 1.3, line on Artificial Intelligence).

\bibliographystyle{plainnat}  
\bibliography{references}

\end{document}